\documentclass[runningheads]{llncs}
\usepackage[T1]{fontenc}
\usepackage{graphicx}%
\usepackage{subcaption}
\usepackage{multirow}%
\usepackage{cite}
\usepackage{amsmath,amssymb,amsfonts}%
\usepackage{booktabs}%
\usepackage{orcidlink}
\usepackage{color}
\hypersetup{
    colorlinks=true,
    linkcolor=blue,
    filecolor=blue,      
    urlcolor=blue,
    citecolor=blue,
}
\begin{document}
\title{Learning Word Embedding with Better Distance Weighting and Window Size Scheduling}
\titlerunning{Learning Word Embedding with LFW and EDWS}
%
%
\author{Chaohao Yang\inst{1} \and Chris Ding\inst{1,2}}
\authorrunning{C. Yang \and C. Ding}
%
%
\institute{School of Data Science, The Chinese University of Hong Kong (Shenzhen)\\Longxiang Avenue, Shenzhen, 518172, China\\ 
\and
Corresponding author(s). E-mail(s): \email{chrisding@cuhk.edu.cn}}
\maketitle              
\begin{abstract}
Distributed word representation (a.k.a. word embedding) is a key focus in natural language processing (NLP). As a highly successful word embedding model, Word2Vec offers an efficient method for learning distributed word representations on large datasets. However, Word2Vec lacks consideration for distances between center and context words. We propose two novel methods, Learnable Formulated Weights (LFW) and Epoch-based Dynamic Window Size (EDWS), to incorporate distance information into two variants of Word2Vec, the Continuous Bag-of-Words (CBOW) model and the Continuous Skip-gram (Skip-gram) model. For CBOW, LFW uses a formula with learnable parameters that best reflects the relationship of influence and distance between words to calculate distance-related weights for average pooling, providing insights for future NLP text modeling research. For Skip-gram, we improve its dynamic window size strategy to introduce distance information in a more balanced way. Experiments prove the effectiveness of LFW and EDWS in enhancing Word2Vec's performance, surpassing previous state-of-the-art methods.
\keywords{Natural language processing \and Word embedding \and Word2Vec \and Learnable weights \and Window size scheduling}
\end{abstract}
\section{Introduction}

NLP researchers have long aimed to obtain high-quality word vector representations. One traditional approach is one-hot encoding, where a vector has a ``1" at the index corresponding to the word and ``0"s elsewhere. However, this approach suffers from the curse of dimensionality when dealing with large vocabulary sizes that can reach millions \cite{bengio2003neural}. Additionally, one-hot encoding fails to capture syntactic or semantic properties because all word distances in the vector space are equal.
    
Distributed word representations have been developed to overcome the limitations of one-hot encoding \cite{hinton1986learning}. In this approach, words are represented by lower-dimensional vectors, typically a few hundred dimensions, where each element can take on various values \cite{schakel2015measuring}. These distributed word vectors can capture both syntactic and semantic properties, allowing syntactic and semantic similarities to be measured using Euclidean distance or cosine similarity between vectors. Distributed word representations may also preserve analogical relationships between words \cite{mikolov2013efficient}. For example, the vector subtraction $vector(``king") - vector(``man") + vector(``woman")$ is most likely to result in a vector closest to $vector(``queen")$. Among all distributed word representation models, the Word2Vec model \cite{mikolov2013efficient} is the most successful one, with outstanding performance in both modeling effectiveness and training efficiency. 

\begin{figure}[t]
    \centering
    \subcaptionbox{}{
        \includegraphics[width = .46\linewidth]{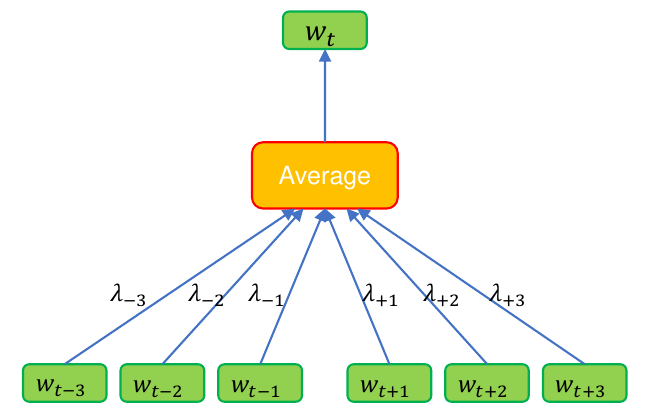}
    }
    \subcaptionbox{}{
    \centering
        \includegraphics[width = .46\linewidth]{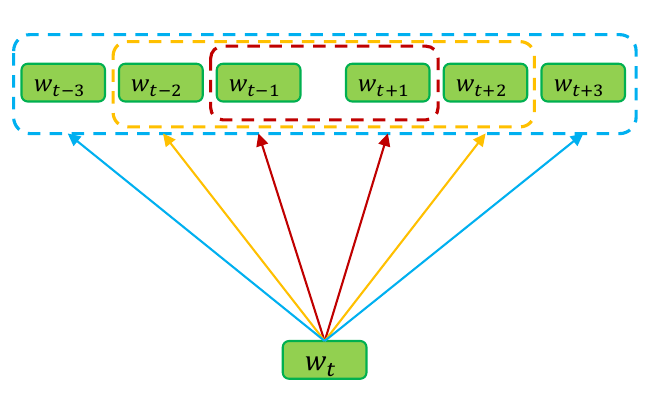}
    }
    \caption{Illustrations of two previous methods for improving Word2Vec with example window size 3. The distance-related weights method in (a) combines all context words ($w_{t+i}$) with their corresponding weights ($\lambda_i$) when averaging them up to predict the center word ($w_t$). The dynamic window size strategy in (b) uses dynamically selected window sizes (indicated by the dashed box in the figure) to sample more from nearby context words, allowing them to contribute more to the training process.}\label{weight&size}
\end{figure}

However, both variants of Word2Vec, namely CBOW and Skip-gram, discard the distance information between center and context words when modeling text \cite{qiu2014learning, chang2017weighted}. Specifically, CBOW predicts the center word (also known as the current word) based on the unweighted average of context word embeddings within a context window around it, while Skip-gram gives the prediction for one context word in the window each time based on the center word. All context words, regardless of their distances to the center word, are treated equally in those predictive tasks. There are two major problems concerning this design: 1) Words closer to each other in distance are usually also closer in their relationships, and their predictive powers with respect to each other are generally larger compared to words that are farther from them, which should be considered in Word2Vec's predictive tasks. 2) The distance between words reflects the order and relative positions of words in a text, which is essential for extracting the meaning of the text. For example, the meanings of ``people can save water'' and ``water can save people'' are greatly different despite the fact that they contain the same words. Therefore, Word2Vec discarding word distance information encounters difficulties in both its predictive training process and semantic modeling ability.

Therefore, many researchers have been working on introducing distance information into these models. For CBOW, some researchers have proposed adding distance-related weights to the context average to make the model distance-sensitive \cite{qiu2014learning, chang2017weighted}. For Skip-gram, exactly one center word and one context word are used each time for modeling and no average is needed. Therefore, The most preferred way among researchers is to dynamically select a random context window size for each center word \cite{mikolov2013efficient} and only use the context words with distances within this dynamic window size for prediction. Since context words closer to the center word are more likely to be in the window, the probability of each context word being used will decrease linearly as their distance to the center word increases, allowing context words closer to the center to contribute more to the training process. Figure~\ref{weight&size} provides some illustrations for these two methods.

Despite their effectiveness, some problems still remain for these distance information introducing methods. For distance-related weights, the major concern is how to construct weights that can utilize a reasonable prior relationship with distances and are adaptive to specific situations. For the dynamic window size strategy, the problem is the frequent and irregular changes in window size caused by random selections, which ruins the training balance for each word and reduces modeling performance.

We propose two novel methods, Learnable Formulated Weights (LFW) and Epoch-based Dynamic Window Size (EDWS), to solve these problems and improve the performance of Word2Vec. LFW introduces a prior formula with a small number of learnable parameters to calculate the distance-related weights. EDWS cancels the random selection while preserving the dynamicity by gradually increasing the window size by the number of training epochs. The effectiveness of both our methods has been demonstrated by various experiments. Specifically, we accomplish an accuracy improvement of 15.3\% with LFW on CBOW and 2.5\% with EDWS on Skip-gram.

\section{Related Work}

The distributed representation of words has a long history since \cite{hinton1986learning} and \cite{elman1991distributed}. In its subsequent development, it gradually accepted the distributional hypothesis presented in distributional representation papers such as Latent Semantic Indexing \cite{blei2003latent} and Latent Dirichlet Allocation \cite{deerwester1990indexing} and derived a series of methods that use context words to obtain the distributed representation of center words. A famous neural network language model (NNLM) architecture for generating word representation of this kind is proposed in \cite{bengio2003neural}, which uses a window of context words to predict the word right after them. In NNLM, context words are first embedded into distributed vector representations, then sequentially concatenated together as input to the following hyperbolic tangent hidden layer, and finally transformed into probability distributions in the vocabulary space at the output layer as the predicted result. In this process, the order of the sequentially concatenated word vectors reflects the order of the context words, which takes the distance information into account.

Thanks to its outstanding modeling ability of contextual relationships, NNLM has been applied in fields like speech recognition and machine translation \cite{devlin2014fast, shi2014efficient} and in research of developing new NLP training methods \cite{turian2010word}. However, NNLM runs too slowly and is difficult to train on large datasets. \cite{schwenk2005training} and \cite{morin2005hierarchical} point out this problem, and many attempts to solve it have been made since then. Among all attempts, \cite{mikolov2013efficient} creatively uses part of the NNLM architecture (mainly removing the hidden layer of NNLM) to train word vectors, which effectively improves the training efficiency and makes the new model called Word2Vec suitable for large datasets. \cite{mikolov2013distributed} also proposes hierarchical softmax, negative sampling, and subsampling methods to further enhance Word2Vec's performance. Since then, Word2Vec has become one of the most popular lightweight techniques for learning word embeddings \cite{karani2018introduction} and is still widely used in recent applications like fake news detection, sentiment analysis, and depression expression classification \cite{mallik2024word2vec, aoumeur2023improving, rakshit2024supervised, nugraha2024classification}.

However, one major issue concerning Word2Vec is that both of its architecture variants, namely CBOW and Skip-gram, discard the concatenation process of the NNLM word embedding layer and its distance-modeling ability. \cite{mikolov2013efficient} proposes the dynamic window size strategy to solve this problem, giving each context word a distance-related sample probability by randomly changing the window size, but at the cost of ruining the training balance for each word. Other studies like \cite{qiu2014learning} and \cite{chang2017weighted} attempt to make CBOW distance-sensitive by introducing distance-related weights. The former provides an independent learnable weight to each context word position, while the latter uses fixed weights calculated by a prescribed distance-related formula. However, These methods either do not model the prior relationships between distance and weights or lose the adaptability of weights. Another approach presented in \cite{ling2015two} introduces different projection layers for different word positions to capture their distance information, which suffers from a much more complex architecture and higher training costs. In summary, further research is still needed to explore more effective methods to incorporate distance information into the Word2Vec model.

\section{Proposed Methods}

\subsection{Learnable Formulated Weights}

In the CBOW model, on a sliding window of the text segment $(w_{t-r}, ..., w_{t-2}, w_{t-1},\\w_{t}, w_{t+1}, w_{t+2}, ..., w_{t+r})$ with window size $r$, we predict the center word $w_t$'s occurrence probability using the context words  $C_t = (w_{t-r}, ..., w_{t-1}, w_{t+1}, ..., w_{t+r})$ as

\begin{equation}
\label{eq1}
P(w_t | C_t ) = \exp(u_{w_t} \cdot u_{C_t}) / \sum_{w' \in V} exp( u_{w'} \cdot u_{C_t} )
\end{equation}

\noindent where $u_{w_t} = E x_{w_t}$, $u_{C_t} = (\sum_{|i|\leq r, i\neq0}u_{t+i})/|C_t|$, and $V$ is the vocabulary set.

Here, $E$ is the neural network weight matrix of size $d \times |V|$, and $x_t$ is the one hot representation of word $w_t$ with element 1 at position $t$ and zero on all other positions. So, $u_{w_t} = E x_{w_t}$ picks the $t$-th column of the $E$ matrix, and this column therefore can be viewed as a distributed representation of the word $w_t$, i.e., the $d$-dimensional embedding of $w_t$.

The above model is not distance-sensitive, since the distance between a context word and the target (center) word is absent in Eq.\ref{eq1}. However, Our experience says that $w_{t+1}$'s relation with $w_t$ is more significant than $w_{t+2}$'s relation with $w_t$. Consequently, $w_{t+1}$'s predictive power  with respect to $w_t$ is generally larger than $w_{t+2}$'s predictive power w.r.t $w_t$.

For this reason, it is reasonable to add a distance-related weight $\lambda_i$ for each context word $u_{t+i}$ to reflect their influence on predicting $w_t$ when averaging them up for $u_{C_t}$. Specifically
\begin{equation}
\label{eq2}
u_{C_t} = \frac{1}{Z}\sum_{|i|\leq r,i\neq0}\lambda_iu_{t+i}
\end{equation}
where $Z=\sum_{|i|\leq r, i\neq0}\lambda_i$ is the normalization factor. 

The main question to be answered for this strategy is how the distance-related weights are constructed. We propose the Learnable Formulated Weights (LFW) method to address this issue, which constructs a formula with a small number of learnable parameters that takes distance as input to calculate weights at different distances. This method combines the prior form of the formula with the posterior learning results of the parameters, allowing us to directly model the relationship between weights and distances and perform dynamic adjustments on specific values of parameters.

There are two major aspects of the prior relationship between weights and distances we will model in this paper, each corresponding to two assumptions. The first aspect is whether or not the two context words on both sides of $w_t$ with the same distance share the same weight, and the second aspect is which kind of mathematical relationship (exponential or power) better models the relationship between weights and distances on a single side. By combining assumption choices for each of the aspects, we will be able to model the prior relationship in four different ways and find out the best combination for CBOW.

Now we will give the formula for each of the combinations of assumptions mentioned above. Note that for each formula there is a learnable constant added to the end to model the basic amount of relation between any two words in the same text to keep them integrated together. The first combination of assumptions is the same weights on both sides and the negative power law decay of weights with respect to distance, and the formula representing this is
\begin{equation}
    \label{eq3}
    \lambda_i=|i|^{-\alpha}+\beta
\end{equation}
\noindent where $0 < |i| \leq r$ represents the distance between a context word and $w_t$, and $\alpha$ and $\beta$ are two learnable parameters initialized to 0 before training.

The second combination of assumptions is different weights on both sides and the negative power law decay of weights with respect to distance, so the formula for this should be
\begin{equation}
    \label{eq4}
    \lambda_i=
    \begin{cases}
        \quad|i|^{-\alpha_0}+\beta_0 \quad \text{if } -r \leq i < 0,\\
        \quad|i|^{-\alpha_1}+\beta_1 \quad \text{if } 0 < i \leq r
    \end{cases}
\end{equation}
\noindent \vspace{4pt}where $\alpha_0$, $\beta_0$, $\alpha_1$, and $\beta_1$ are four learnable parameters initialized to 0 before training.

Based on the content above, the formulas corresponding to the third and fourth assumption combinations can be given accordingly as Eq.\ref{eq5} and Eq.\ref{eq6}, except that the weights decrease exponentially with respect to distance:
\begin{equation}
    \label{eq5}
    \lambda_i=e^{-\alpha|i|}+\beta
\end{equation}
\begin{equation}
    \label{eq6}
    \lambda_i=
    \begin{cases}
        \quad e^{-\alpha_0|i|}+\beta_0 \quad \text{if } -r \leq i < 0,\\
        \quad e^{-\alpha_1|i|}+\beta_1 \quad \text{if } 0 < i \leq r
    \end{cases}
\end{equation}

In the experimental part, we will use the above four formulas for weight calculation respectively to demonstrate which of those formulas is the most effective, and at the same time prove the superiority of our method over other approaches to adding weights to CBOW.

\subsection{Epoch-based Dynamic Window Size}

For the Skip-gram model, each center word $w_t$ is used as an input to predict context words in a sliding window of size $r$. That is, for each context word $w_{t+i}$ with $0 < |i| \leq r$, we predict its occurrence probability as

\begin{equation}
    \label{eq7}
    P(w_{t+i}|w_t) = exp(u_{w_{t+i}}\cdot u_{w_t})/\sum_{w'\in V} exp(u_{w'}\cdot u_{w_t})
\end{equation}

Because no average of context words appears in this model, we cannot make the model distance-sensitive by adding weights for context words. However, since the context words that appear more in the prediction have more influence on how the word embeddings are trained, we can make the model sensitive to distance by sampling more from context words that are closer to the center word and sampling less otherwise. In other words, the probability of being sampled will act in a similar way as the weights added to the CBOW model.

A naive way to do this is to dynamically select a random context window size $r'$ from 1 to $r$ for each center word $w_t$, and use $w_t$ to predict only context words within distance $r'$. In this case, the probability of each $w_{t+i}$ with $0 < |i| \leq r$ to be used for prediction is

\begin{equation}
    \label{eq8}
    P(w_{t+i}) = P(|i| \leq r') = \frac{r-|i|+1}{r}
\end{equation}

We can learn from Eq.\ref{eq8} that the probability of each $w_{t+i}$ to be used for prediction decreases linearly as the distance $|i|$ increases. Therefore, the context words closer to $w_t$ will obtain larger weights than those farther away.

One major problem concerning this dynamic window size strategy is the frequent and irregular changes in window size for each center word resulting from random selections, which ruins
the training balance for each word and reduces modeling performance. For example, imagine that the window sizes selected for two center words $w_{t_1}$ and $w_{t_2}$ are 1 and r respectively, then the number of predictions made by $w_{t_2}$ is $r$ times larger than that of $w_{t_2}$, causing the model to pay special attention to the center words in some parts while ignoring the training of others, which will lead to an unreasonable allocation of calculations and a decrease in model performance.

We propose the Epoch-based Dynamic Window Size (EDWS) to solve this problem, which cancels the random selection while preserving the dynamicity by gradually increasing the window size by the number of training epochs. In our specific implementation, we use three different window sizes with a ratio of $1:2:3$ to represent the beginning, middle, and end of the training process. The smallest window size appears first, then the middle and largest one, each running for the same number of epochs. Therefore, the window size $r'_k$ on the $k$th epoch is given by

\begin{equation}
    \label{eq9}
    r'_k = \lceil \frac{3k}{K} \rceil \frac{r}{3}
\end{equation}

\noindent where $K$ is the total number of epochs and r is the maximum window size, both are multiples of 3 in this case, but can be other values if under a different implementation.

The EDWS strategy has two major advantages over the original dynamic window size. First, EDWS eliminates the irregularity brought by random selections at its source. Second, EDWS specifies the window sizes to appear in ascending order, forming a gradual learning process from nearby to distant contexts, which aligns with general learning principles. Given these advantages, EDWS is theoretically more effective than the original dynamic window size, which will be demonstrated through experiments in later parts of this paper.

\section{Experiments}

\renewcommand{\arraystretch}{1.1}
\begin{table}[t]
    \centering
    
    \begin{tabular}{ccccc}
        \toprule
        Model                      & Window Size & Semantic & Syntactic & Total \\ \midrule
        \multirow{5}{*}{CBOW}      & 5                   & 960      & 2987      & 3947  \\ \cmidrule{2-5} 
                                   & 10                  & 1317     & 2938      & 4255  \\ \cmidrule{2-5} 
                                   & 15                  & 1687     & 2814      & 4501  \\ \cmidrule{2-5} 
                                   & 20                  & 1756     & 2867      & 4623  \\ \midrule
        \multirow{5}{*}{Skip-gram} & 5                   & 1995     & 2590      & 4585  \\ \cmidrule{2-5} 
                                   & 10                  & 2179     & 2673      & 4852  \\ \cmidrule{2-5} 
                                   & 15                  & 2666     & 2368      & 5034  \\ \cmidrule{2-5} 
                                   & 20                  & 2710     & 2429      & 5139  \\ \bottomrule
    \end{tabular}
    
    \vspace{4pt}\caption{Performance of CBOW and Skip-gram with different window sizes. Because some (2392 out of 19544) analogical reasoning questions involve words that do not appear in the vocabulary of the training dataset, the number of correct answers is reported rather than the accuracy.}\label{table1}
\end{table}

In this section, we first find the appropriate maximum window size via multiple pre-experiments. Then, we test the effectiveness of LFW on CBOW and EDWS on Skip-gram, each compared with its original model.

\subsection{Datasets}

We use two corpora constructed from English Wikipedia for training: enwik9 and text8 \footnote{Datasets and text preprocessing script are from \url{http://mattmahoney.net/dc/textdata.html}}. The enwik9 dataset contains the first $10^9$ bytes of English Wikipedia with about 120 million words after being processed by Matt Mahoney’s text preprocessing script, and text8 contains the first $10^8$ bytes of the preprocessed enwik9 with about 17 million words. We discard words that appear no more than 5 times in each dataset to focus on more valuable words while reducing computation overhead, forming two vocabularies with sizes of 194,377 and 63,643. The enwik9 dataset is much larger than text8, so we conduct pre-experiments on text8 to determine the hyperparameter choice, i.e., the window size. Then, we conduct comparative experiments on enwik9 based on the chosen hyperparameter to demonstrate the effectiveness of our proposed methods.

As for test results, We use the analogical reasoning task dataset proposed in \cite{mikolov2013efficient} to measure the quality of the trained word embeddings. By evaluating the embeddings' analogical inference ability, this task reflects the embeddings' mastery of language and forces a more precise distribution of the trained word embeddings in the vector space to meet the requirements of linguistic relations. There are 8,869 semantic and 10,675 syntactic questions in the test set in total. The semantic questions are divided into five categories, while the syntactic questions are divided into nine categories, as shown in Table~\ref{table4}. An example of these questions is finding the word with the same relative relationship with the word ``woman'' as the word ``man'' has with the word ``king''. The question should be answered by finding the word embedding that is closest to $vector(``king") - vector(``man") + vector(``woman")$ in cosine distance. The quality of word embeddings is then measured by the total accuracy of all questions in this task.

\subsection{Window Size Choice}

We list the results of CBOW and Skip-gram with different window sizes in Table~\ref{table1}. Since the dynamic window size strategy is proposed as an original design of Skip-gram, we will use this strategy for all original Skip-gram models used in the experiments, and the window size represents its maximum window size in random selection. We set the word vector dimension to a relatively small value of 128 in this step to reduce the calculation cost of pre-experiments. Consistent with the results reported in \cite{chang2017weighted}, we find that the improvement in model performance is significantly related to the increase in window size before the window size reaches 15. After that, the correlation becomes less significant. Considering the impact of increasing the window size on the calculation amount, we choose the window size to be 15 for formal experiments.

\subsection{Experimental Results}

\begin{table}[t]
    \centering
    \begin{tabular}{cccc}
        \toprule
        Model          & Semantic            & Syntactic            & Total                 \\ \midrule
CBOW           & 27.26\% & 45.70\% & 37.34\%  \\ \midrule
LFW CBOW Eq.\ref{eq3}        & \textbf{48.39}\% & \textbf{56.24\%} & \textbf{52.68}\% \\ \midrule
LFW CBOW Eq.\ref{eq4}        & 47.32\% & 55.46\% & 51.77\% \\ \midrule
LFW CBOW Eq.\ref{eq5}        & 44.51\% & 55.77\% & 50.66\% \\ \midrule
LFW CBOW Eq.\ref{eq6}        & 43.14\% & 54.75\% & 49.48\% \\ \bottomrule
    \end{tabular}
    \vspace{4pt}\caption{Performance of LFW}\label{table2}
\end{table}

\begin{table}[t]
    \centering
    \begin{tabular}{cccc}
        \toprule
        Model          & Semantic            & Syntactic            & Total                 \\ \midrule
Skip-gram      & 58.43\% & 29.63\% & 42.70\%  \\ \midrule
EDWS Skip-gram & \textbf{61.60\%} & \textbf{31.59}\% & \textbf{45.21}\%  \\ \bottomrule
    \end{tabular}
    \vspace{4pt}\caption{Performance of EDWS}\label{table3}
\end{table}

In this step, we conduct experiments to show our methods' effectiveness for Word2Vec on the enwik9 dataset. We set the word vector dimension to 600, following the configuration in \cite{qiu2014learning}. Table~\ref{table2} and Table~\ref{table3} show positive experimental results for both proposed methods, where the fonts in bold indicate the highest accuracy for each column. The LFW methods following Eq.\ref{eq3}, Eq.\ref{eq4}, Eq.\ref{eq5}, and Eq.\ref{eq6} improve the overall accuracy of CBOW by 15.3\%, 14.4\%, 13.3\%, and 12.1\%, respectively, and EDWS improves that of Skip-gram by more than 2.5\%.

\begin{figure}[t]
    \centering
    \includegraphics[width=99mm, height=76mm]{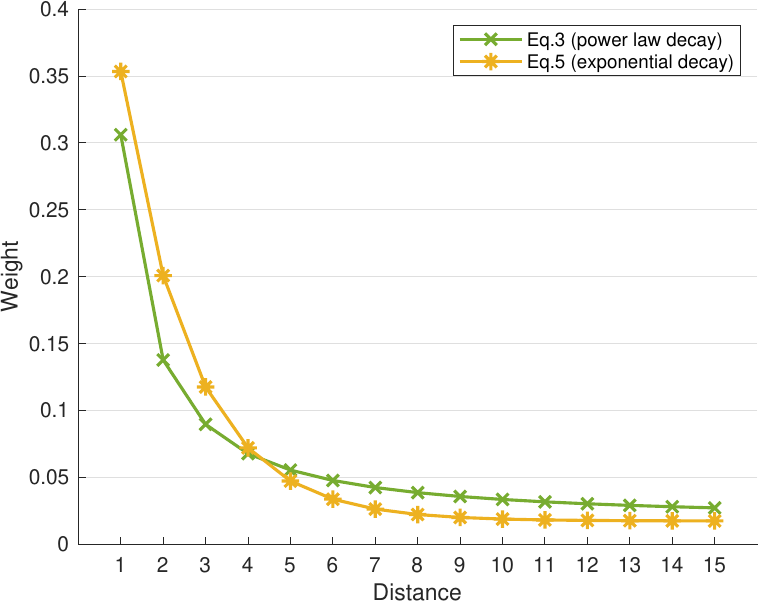}
    \caption{The curves of normalized weights for power law decay and exponential decay}\label{new all}
\end{figure}

Results suggest that formula Eq.\ref{eq3} models the relationship between weight and distance best among all proposed formulas. This may be because: 1) There is no significant difference in the relationship between context on two sides and the center word and using the same parameters on both sides can reduce overfitting. 2) The closest context words have a significantly higher influence on the center word than the other, while the influence of the other words decreases relatively slowly with distance, which can be modeled better with the power function that decreases greatly from distance 1 to 2 and gently at farther distances. To establish an intuitive impression of the relationship between weights and distances obtained using power and exponential functions, we provide Figure~\ref{new all}, which shows the curves formed by the relationship between weights and distances learned from the enwik9 dataset when using Eq.\ref{eq3} (power law decay) and Eq.\ref{eq5} (exponential decay). The curves in the figure take their points at integer distances within the window size, and the sum of the weights for each curve is normalized to 1 for ease of comparison. The behavior of those curves solidifies our analysis of the success of Eq.\ref{eq3}. Therefore, we suggest using Eq.\ref{eq3} as the formula for LFW and future context modeling efforts. to achieve the largest performance improvement.

Under their best version, Our methods' performance exceeds the state-of-the-art accuracy improvement of 13.5\% on CBOW and 1.8\% on Skip-gram achieved by independent learnable weights presented in \cite{qiu2014learning}. Additionally, the LFW method consumes less than 15\% of extra training time, which is a huge improvement compared with the 72\% (computed from its 42\% slower speed) consumed by the method in \cite{qiu2014learning}. The test results and the improvement amounts of LFW Eq.\ref{eq3} (best version) detailed to each category are shown in Table~\ref{table4} to illustrate its effectiveness better. Significant relative improvements are brought to nearly every category by our proposed method. The experimental results prove the effectiveness of both our methods.

\begin{table}[t]
    \begin{tabular}{cccc}
    \toprule
    Categories                  & CBOW                 & LFW CBOW Eq.\ref{eq3}         & $\Delta$ Acc (\%) \\ \midrule
    \textbf{Semantic}        & 27.26\% (2418/8869)  & 48.39\% (4292/8869)   & 21.13\% (77.50\%)
           \\ \midrule
    capital-common-countries & 80.04\% (405/506)    & 92.49\% (468/506)     & 12.45\% (15.56\%)          \\ 
    capital-world            & 31.06\% (1405/4524)  & 62.07\% (2808/4524)   & 31.01\% (99.86\%)          \\ 
    currency                 & 3.00\% (26/866)      & 3.93\% (34/866)       & 0.92\% (30.77\%)           \\ 
    city-in-state            & 15.08\% (372/2467)   & 28.70\% (708/2467)    & 13.62\% (90.32\%)          \\ 
    family                   & 41.50\% (210/506)    & 54.15\% (274/506)     & 12.65\% (30.48\%)
          \\ \midrule
    \textbf{Syntactic}       & 45.70\% (4879/10675) & 56.24\% (6004/10675)  & 10.54\% (23.06\%)          \\ \midrule
    adjective-to-adverb      & 11.49\% (114/992)    & 17.34\% (172/992)     & 5.85\% (50.88\%)           \\ 
    opposite                 & 17.49\% (142/812)    & 31.03\% (252/812)     & 13.55\% (77.46\%)          \\ 
    comparative              & 65.62\% (874/1332)   & 76.73\% (1022/1332)   & 11.11\% (16.93\%)          \\ 
    superlative              & 21.75\% (244/1122)   & 38.86\% (436/1122)    & 17.11\% (78.69\%)          \\ 
    present-participle       & 44.22\% (467/1056)   & 45.17\% (477/1056)    & 0.95\% (2.14\%)            \\ 
    nationality-adjective    & 75.05\% (1200/1599)  & 84.24\% (1347/1599)   & 9.19\% (12.25\%)           \\ 
    past-tense               & 41.60\% (649/1560)   & 45.38\% (708/1560)    & 3.78\% (9.09\%)           \\ 
    plural                   & 62.31\% (830/1332)   & 81.68\% (1088/1332)   & 19.37\% (31.08\%)          \\ 
    plural-verbs             & 41.26\% (359/870)    & 57.70\% (502/870)     & 16.44\% (39.83\%)          \\ \midrule
    \textbf{Total}           & 37.34\% (7297/19544) & 52.68\% (10296/19544) & 15.34\% (41.10\%)          \\ \bottomrule
    \end{tabular}
    \vspace{4pt}\caption{Performance of LFW Eq.\ref{eq3} on each category of the analogical reasoning task. $\Delta$ indicates the absolute and relative performance improvement of LFW Eq.\ref{eq3} compared to original CBOW.}\label{table4}
\end{table}

\section{Conclusion}

In this paper, we propose two novel methods, LFW and EDWS, to improve the performance of Word2Vec. For CBOW, We introduce distance-related weights calculated by a prior formula with a small number of learnable parameters, the best version of which can be widely adopted in NLP text modeling tasks and research regarding how the influence and relevance between words change with distance. For Skip-gram, we invent an epoch-based dynamic window size method to eliminate the irregularity brought by dynamic window size at its source. Experiments prove that both methods effectively improve the performance of their corresponding models, surpassing state-of-the-art approaches to incorporating distance information into Word2Vec. For future work, we will focus on combining the two methods to improve model performance, such as specifying the running epochs of different window sizes based on the learnable formulated weights at different distances. Research related to this area should be a valuable direction in the future.







\bibliographystyle{splncs04}
\bibliography{mybib}
\end{document}